\documentclass[10pt,twocolumn,letterpaper]{article}

\usepackage{iccv}
\usepackage{times}
\usepackage{epsfig}
\usepackage{graphicx}
\usepackage{amsmath}
\usepackage{amssymb}

\usepackage{enumitem} 
\usepackage{bbm}
\usepackage{caption}
\usepackage{subcaption}
\usepackage{url}

\usepackage{array}
\usepackage{multirow}
\usepackage{graphicx}
\usepackage{boldline}
\usepackage{booktabs}

\usepackage{textcomp}
\usepackage{siunitx}

\makeatletter
\@namedef{ver@eso-pic.sty}{}
\makeatother
\usepackage{pdfpages}
\usepackage[pagebackref=true,breaklinks=true,letterpaper=true,colorlinks,bookmarks=false]{hyperref}

\iccvfinalcopy 

\def\iccvPaperID{8167} 

\ificcvfinal\fi

\begin{document}

\title{Exploiting a Joint Embedding Space for Generalized Zero-Shot Semantic Segmentation}

\author{Donghyeon Baek\thanks{Equal contribution,~$^\dagger$Corresponding author.}
\quad\quad\quad
Youngmin Oh\footnotemark[1]
\quad\quad\quad
Bumsub Ham\footnotemark[2]
\\
School of Electrical and Electronic Engineering, Yonsei University
\\
\url{https://cvlab.yonsei.ac.kr/projects/JoEm}
}

\maketitle
\ificcvfinal\thispagestyle{empty}\fi

\begin{abstract}
We address the problem of generalized zero-shot semantic segmentation~(GZS3) predicting pixel-wise semantic labels for seen and unseen classes. Most GZS3 methods adopt a generative approach that synthesizes visual features of unseen classes from corresponding semantic ones~(e.g.,~\textit{word2vec}) to train novel classifiers for both seen and unseen classes. Although generative methods show decent performance, they have two limitations: (1) the visual features are biased towards seen classes; (2) the classifier should be retrained whenever novel unseen classes appear. We propose a discriminative approach to address these limitations in a unified framework. To this end, we leverage visual and semantic encoders to learn a joint embedding space, where the semantic encoder transforms semantic features to semantic prototypes that act as centers for visual features of corresponding classes. Specifically, we introduce boundary-aware regression~(BAR) and semantic consistency~(SC) losses to learn discriminative features. Our approach to exploiting the joint embedding space, together with BAR and SC terms, alleviates the seen bias problem. At test time, we avoid the retraining process by exploiting semantic prototypes as a nearest-neighbor~(NN) classifier. To further alleviate the bias problem, we also propose an inference technique, dubbed Apollonius calibration~(AC), that modulates the decision boundary of the NN classifier to the Apollonius circle adaptively. Experimental results demonstrate the effectiveness of our framework, achieving a new state of the art on standard benchmarks.
\end{abstract}

\section{Introduction}\vspace{-.2cm}
\begin{figure}[t]
    \centering
    \begin{subfigure}{\linewidth}
        \centering
        \includegraphics[width=\linewidth]{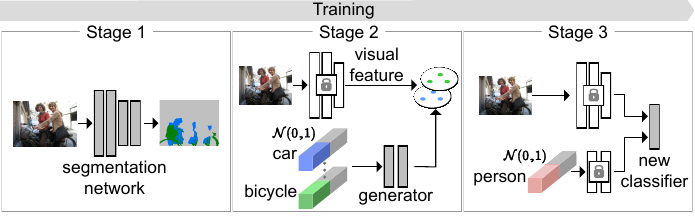}
    \end{subfigure}
    \begin{subfigure}{\linewidth}
        \centering
        \includegraphics[width=\linewidth]{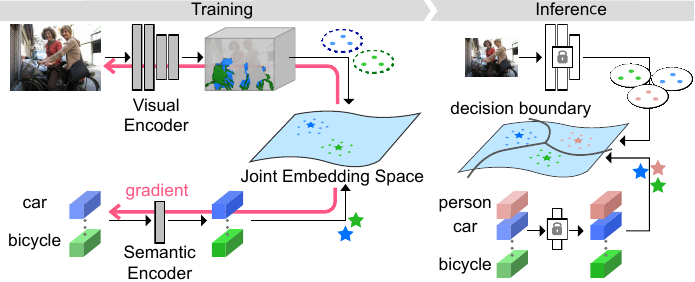}
    \end{subfigure}
    \vspace{-.2cm}
    \caption{\small{In contrast to generative methods~\cite{ZS3NET19Bucher,CSRL20Li}~(top), we update both visual and semantic encoders to learn a joint embedding space, and leverage a nearest neighbor classifier in the joint embedding space at test time~(bottom). This alleviates a bias problem towards seen classes, and avoids re-training the classifier. We visualize visual features and semantic prototypes by circles and stars, respectively. Best viewed in color.}}
    \vspace{-.5cm}
    \label{fig:fig1}
\end{figure}
Recent works using convolutional neural networks~(CNNs)~\cite{V3PLUS18Chen,FCN15Long,UNet15Ronne,PSPNet17Zhao} have achieved significant success in semantic segmentation. They have proven effective in various applications such as image editing~\cite{IMAGE18Liang} and autonomous driving~\cite{AUTO20Yurt}, but semantic segmentation in the wild still has two limitations. First, existing methods fail to generalize to new domains/classes, assuming that training and test samples share the same distribution. Second, they require lots of training samples with pixel-level ground-truth labels prohibitively expensive to annotate. As a result, current methods could handle a small set of pre-defined classes only~\cite{SINGLEGPU20Jain}.

As alternatives to pixel-level annotations, weakly-supervised semantic segmentation methods propose to exploit image-level labels~\cite{SeeNet18Hou}, scribbles~\cite{ScribbleSup16Lin}, and bounding boxes~\cite{BoxSup15Dai}, all of which are less labor-intensive to annotate. These methods, however, also require a large number of weak supervisory signals to train networks for novel classes. On the contrary, humans can easily learn to recognize new concepts in a scene with a few visual examples, or even with descriptions of them. Motivated by this, few- and zero-shot learning methods~\cite{ZSL08Laro,ZSL09Pala,MatchNet16Vinyals} have been proposed to recognize objects of previously unseen classes with a few annotated examples and even without them, respectively. For example, few-shot semantic segmentation~(FS3) methods~\cite{OSLSM17Shaban,PANet19Wang} typically exploit an episode training strategy, where each episode consists of randomly sampled support and query sets, to estimate query masks with a few annotated support examples. Although these FS3 methods show decent performance for unseen classes, they are capable of handling a single unseen class only. Recently, the work of~\cite{VOCA17Zhao} first explores the problem of zero-shot semantic segmentation~(ZS3), where it instead exploits pre-trained \textit{semantic features} using class names~(\emph{i.e.},~\textit{word2vec}~\cite{W2V13Mikolov}). This work, however, focuses on predicting unseen classes, even if a given image contains both seen and unseen ones. To overcome this, generalized ZS3~(GZS3) has recently been introduced to consider both seen and unseen classes in a scene during inference. Motivated by generative approaches~\cite{CROSSVALID117Bucher,ZSLGEN18Wang,ZSLGEN18Xian} in zero-shot image classification, many GZS3 methods~\cite{ZS3NET19Bucher,CAGNET20Gu,CSRL20Li} first train a segmentation network that consists of a feature extractor and a classifier with seen classes. They then freeze the feature extractor to extract \textit{visual features}, and discard the classifier. With the fixed feature extractor, a generator~\cite{GAN14Good,VAE13Kingma} is trained to produce visual features from semantic ones (\emph{e.g.},~\textit{word2vec}) of corresponding classes. This enables training novel classifiers with real visual features of seen classes and generated ones of unseen classes~(Fig.~\ref{fig:fig1} top). Although generative methods achieve state-of-the-art performance in GZS3, they have the following limitations:~(1) the feature extractor is trained without considering semantic features, causing a bias towards seen classes. The seen bias problem becomes even worse through a multi-stage training strategy, where the generator and novel classifiers are trained using the feature extractor;~(2) the classifier needs to be re-trained whenever a particular unseen class is newly included/excluded, hindering deployment in a practical setting, where unseen classes are consistently emerging.

We introduce a discriminative approach for GZS3, dubbed JoEm, that addresses the limitations of generative methods in a unified framework ~(Fig.~\ref{fig:fig1} bottom). Specifically, we exploit visual and semantic encoders to learn a joint embedding space. The semantic encoder transforms semantic features into semantic prototypes acting as centers for visual features of corresponding classes. Our approach to using the joint embedding space avoids the multi-stage training, and thus alleviates the seen bias problem. To this end, we propose to minimize the distances between visual features and corresponding semantic prototypes in the joint embedding space. We have found that visual features at object boundaries could contain a mixture of different semantic information due to the large receptive field of deep CNNs. Directly minimizing the distances between the visual features and semantic prototypes might distract discriminative feature learning. To address this, we propose a boundary-aware regression~(BAR) loss that exploits semantic prototypes linearly interpolated to gather the visual features at object boundaries along with its efficient implementation. We also propose to use a semantic consistency~(SC) loss that transfers relations between seen classes from a semantic embedding space to the joint one, regularizing the distances between semantic prototypes of seen classes explicitly. At test time, instead of re-training the classifier as in the generative methods~\cite{ZS3NET19Bucher,CAGNET20Gu,CSRL20Li}, our approach to learning discriminative semantic prototypes enables using a nearest neighbor~(NN) classifier~\cite{NN67Cover} in the joint embedding space. In particular, we modulate the decision boundary of the NN classifier using the Apollonius circle. This Apollonius calibration~(AC) method also makes the NN classifier less susceptible to the seen bias problem. We empirically demonstrate the effectiveness of our framework on standard GZS3 benchmarks~\cite{VOC10Ever,CONTEXT14Motta}, and show that AC boosts the performance significantly. The main contributions of our work can be summarized as follows:
\begin{itemize}[leftmargin=*]
    \vspace{-.2cm}
    \item[$\bullet$] We introduce a simple yet effective discriminative approach for GZS3. We propose BAR and SC losses, which are complementary to each other, to better learn discriminative representations in the joint embedding space.
    \vspace{-.2cm}
    \item[$\bullet$] We present an effective inference technique that modulates the decision boundary of the NN classifier adaptively using the Apollonius circle. This alleviates the seen bias problem significantly, even without re-training the classifier.
    \vspace{-.2cm}
    \item[$\bullet$] We demonstrate the effectiveness of our approach exploiting the joint embedding space on standard benchmarks for~GZS3~\cite{VOC10Ever,CONTEXT14Motta}, and show an extensive analysis with ablation studies.
\end{itemize}

\section{Related work}
\vspace{-.15cm}
\noindent \textbf{Zero-shot image classification.}
Many zero-shot learning~(ZSL)~\cite{ZSLATT09Farhadi,ZSL08Laro,ZSL09Pala} methods have been proposed for image classification. They typically rely on side information, such as attributes~\cite{ZSLATT09Farhadi,ZSLATT09Lampert}, semantic features from class names~\cite{ZSLJOINT16Lu,ZSL17Zhang}, or text descriptions~\cite{ZSLJOINT15Lei,ZSLTEXT16Reed}, for relating unseen and seen object classes. Early ZSL methods~\cite{ZSL15Akata,DEVISE13Frome,ZSLTEXT16Reed, ZSL17Zhang} focus on improving performance for unseen object classes, and typically adopt a discriminative approach to learn a compatibility function between visual and semantic embedding spaces. Among them, the works of~\cite{ZSLJOINT14Fu,ZSLJOINT15Lei,ZSLJOINT16Lu,ZSLJOINT15Yang} exploit a joint embedding space to better align visual and semantic features. Similarly, our approach leverages the joint embedding space, but differs in that (1)~we tackle the task of GZS3, which is much more challenging than image classification, and~(2) we propose two complementary losses together with an effective inference technique, enabling learning better representations and alleviating a bias towards seen classes. Note that a straightforward adaptation of discriminative ZSL methods~\cite{ZSL16Chang,ZSLATT13Lamp,ZSL13Norou} to generalized ZSL~(GZSL) suffers from the seen bias problem severely. To address this, a calibrated stacking method~\cite{CALIB16Chaqo} proposes to penalize scores of seen object classes at test time. This is similar to our AC in that both aim at reducing the seen bias problem at test time. The calibrated stacking method, however, shifts the decision boundary with a constant value, while we modulate the decision boundary adaptively. Recently, instead of learning the compatibility function between visual and semantic embedding spaces, generative methods~\cite{CROSSVALID117Bucher,ZSLGEN18Huang,ZSLGEN19Li,ZSLGEN19Sari,ZSLGEN18Wang,ZSLGEN18Xian} attempt to address the task of GZSL by using generative adversarial networks~\cite{GAN14Good} or variational auto-encoders~\cite{VAE13Kingma}. They first train a generator to synthesize visual features from corresponding semantic ones or attributes. The generator then produces visual features of given unseen classes, and uses them to train a new classifier for both seen and unseen classes. In this way, generative methods reformulate the task of GZSL as a standard classification problem, outperforming the discriminative ones, especially on the generalized setting.

\noindent \textbf{Zero-shot semantic segmentation.}
Recently, there are many attempts to extend ZSL methods for image classification to the task of semantic segmentation. They can be categorized into discriminative and generative methods. The work of~\cite{VOCA17Zhao} adopts the discriminative approach for ZS3, focusing on predicting unseen classes in a hierarchical way using WordNet~\cite{WORDNET95Miller}. The work of~\cite{UNCER20Hou} argues that adverse effects from noisy samples are significant especially in the problem of ZS3, and proposes uncertainty-aware losses~\cite{UNCER17Kendall} to prevent a segmentation network from overfitting to them. This work, however, requires additional parameters to estimate the uncertainty, and outputs a binary mask for a given class only. SPNet~\cite{SPNET19Xian} exploits a semantic embedding space to tackle the task of GZS3, mapping visual features to fixed semantic ones. Differently, we propose to use a joint embedding space, better aligning visual and semantic spaces, together with two complementary losses. In contrast to discriminative methods, ZS3Net~\cite{ZS3NET19Bucher} leverages a generative moment matching network~(GMMN)~\cite{GMMN15Li} to synthesize visual features from corresponding semantic ones. Training ZS3Net requires three stages for a segmentation network, the GMMN, and a new classifier, respectively. While ZS3Net exploits semantic features of unseen classes at the last stage only, CSRL~\cite{CSRL20Li} incorporates them in the second stage, encouraging synthesized visual features to preserve relations between seen and unseen classes in the semantic embedding space. CaGNet~\cite{CAGNET20Gu} proposes a contextual module using dilated convolutional layers~\cite{Dilated14Papa} along with a channel-wise attention mechanism~\cite{SENet18Hu}. This encourages the generator to better capture the diversity of visual features. The generative methods~\cite{ZS3NET19Bucher,CAGNET20Gu,CSRL20Li} share the common limitations as follows: First, they require re-training the classifier whenever novel unseen classes are incoming. Second, they rely on the multi-stage training framework, which might deteriorate the seen bias problem, with several hyperparameters (\emph{e.g.}, the number of synthesized visual features and the number of iterations for training a new classifier). To address these limitations, we advocate using a discriminative approach that avoids the multi-stage training scheme and re-training the classifier.

\section{Method}
\vspace{-.15cm}
\begin{figure*}[t]	
	\centering
    \begin{subfigure}{.59\textwidth}
        \centering
        \includegraphics[width=\textwidth]{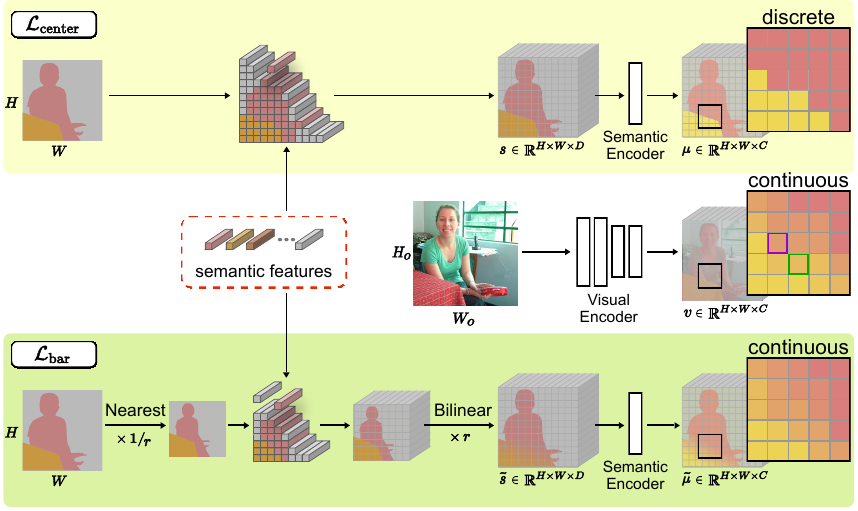}
        \caption{\small{Discrepancy between visual feature and semantic prototype maps.}}
	\end{subfigure}
    \begin{subfigure}{.4\textwidth}
        \centering
		\includegraphics[width=\textwidth]{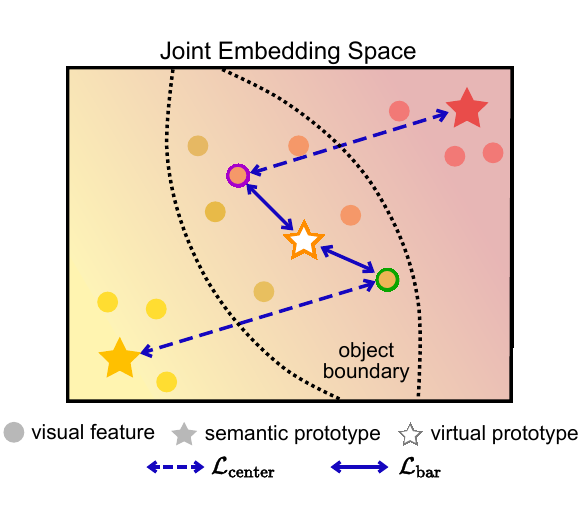}
        \caption{\small{Comparison of~$\mathcal{L}_\mathrm{center}$ and~$\mathcal{L}_\mathrm{bar}$.}}
    \end{subfigure}
    \vspace{-.2cm}
    \caption{\small{(a) While the semantic feature map abruptly changes at object boundaries due to the stacking operation using a ground-truth mask~(top), the visual one smoothly varies due to the large receptive field of the visual encoder~(middle). We leverage a series of nearest-neighbor and bilinear interpolations to smooth a sharp transition at object boundaries in an efficient way~(bottom).~(b) Visual features at object boundaries might contain a mixture of different semantics, suggesting that minimizing the distances to the exact semantic prototypes is not straightforward~(dashed lines). Our BAR loss exploits a virtual prototype to pull the visual features at object boundaries~(solid lines). Best viewed in color.}}
    \label{fig:fig2}
    \vspace{-.45cm}
\end{figure*}
In this section, we concisely describe our approach to exploiting a joint embedding space for GZS3~(Sec.~\ref{sec:sec3_1}), and introduce three training losses~(Sec.~\ref{sec:sec3_2}). We then describe our inference technique~(Sec.~\ref{sec:sec3_3}).

\subsection{Overview} \label{sec:sec3_1}
\vspace{-.1cm}Following the common practice in~\cite{ZS3NET19Bucher,CAGNET20Gu,CSRL20Li,SPNET19Xian}, we divide classes into two disjoint sets, where we denote by~$\mathcal{S}$ and~$\mathcal{U}$ sets of seen and unseen classes, respectively. We train our model including visual and semantic encoders with the seen classes~$\mathcal{S}$ only, and use the model to predict pixel-wise semantic labels of a scene for both seen and unseen classes, $\mathcal{S}$ and~$\mathcal{U}$, at test time. To this end, we jointly update both encoders to learn a joint embedding space. Specifically, we first extract visual features using the visual encoder. We then input semantic features~(\emph{e.g.},~\textit{word2vec}~\cite{W2V13Mikolov}) to the semantic encoder, and obtain semantic prototypes that represent centers for visual features of corresponding classes. We have empirically found that visual features at object boundaries could contain a mixture of different semantics~(Fig.~\ref{fig:fig2}(a) middle), which causes discrepancies between visual features and semantic prototypes. To address this, we propose to use linearly interpolated semantic prototypes~(Fig.~\ref{fig:fig2}(a) bottom), and minimize the distances between the visual features and semantic prototypes~(Fig.~\ref{fig:fig2}(b)). We also encourage the relationships between semantic prototypes to be similar to those between semantic features explicitly~(Fig.~\ref{fig:fig3}). At test time, we use the semantic prototypes of both seen and unseen classes as a NN classifier without re-training. To further reduce the seen bias problem, we modulate the decision boundary of the NN classifier adaptively~(Fig.~\ref{fig:fig4}(c)). In the following, we describe our framework in detail.

\subsection{Training} \label{sec:sec3_2}
\vspace{-.1cm}We define an overall objective for training our model end-to-end as follows:
\begin{equation}
    \mathcal{L} = \mathcal{L}_\mathrm{ce} + \mathcal{L}_\mathrm{bar} + \lambda \mathcal{L}_\mathrm{sc},
\end{equation}
where we denote by~$\mathcal{L}_\mathrm{ce}$,~$\mathcal{L}_\mathrm{bar}$, and~$\mathcal{L}_\mathrm{sc}$ cross-entropy~(CE), BAR, and SC terms, respectively, balanced by the parameter~$\lambda$. In the following, we describe each loss in detail.

\noindent \textbf{CE loss.}
Given an image of size~$H_o \times W_o$, the visual encoder outputs a visual feature map~$v \in \mathbb{R}^{H \times W \times C}$, where~$H$,~$W$,~and~$C$ are height, width, and the number of channels, respectively. We denote by~$y$ a corresponding ground-truth mask, which is resized to the size of~$H \times W$ using nearest-neighbor interpolation, and~$v(\mathbf{p})$ a~$C$-dimensional local visual feature at position~$\mathbf{p}$. To encourage these visual features to better capture rich semantics specific to the task of semantic segmentation, we use a CE loss widely adopted in supervised semantic segmentation. Differently, we apply this for a set of seen classes~(\emph{i.e.},~$\mathcal{S}$) only as follows:
\begin{equation}
    \mathcal{L}_\mathrm{ce} = - \frac{1}{\sum_{c \in \mathcal{S}} |\mathcal{R}_c|} \sum_{c \in \mathcal{S}} \sum_{\mathbf{p}\in \mathcal{R}_c} \log \frac{e^{w_c \cdot v(\mathbf{p})}}{\sum_{j \in \mathcal{S}} e^{w_j \cdot v(\mathbf{p})}},
\end{equation}
where $w_c$ is a~$C$-dimensional classifier weight for a class~$c$ and~$\mathcal{R}_c$ indicates a set of locations labeled as the class~$c$ in~$y$. We denote by~$|\cdot|$ the cardinality of a set.

\noindent \textbf{BAR loss.}
Although the CE loss trains the classifier to discriminate seen classes, the learned classifier weights~$w$ are not adaptable to recognize unseen ones. To address this, we instead use the semantic encoder as a hypernetwork~\cite{HyperNet17Ha} that generates classifier weights. Specifically, the semantic encoder transforms a semantic feature~(\emph{e.g.},~\textit{word2vec}~\cite{W2V13Mikolov}) into a semantic prototype that acts as a center for visual features of a corresponding class. We then use semantic prototypes of both seen and unseen classes as a NN classifier at test time.

A straightforward way to implement this is to minimize the distances between visual features and corresponding semantic prototypes during training. To this end, we first obtain a semantic feature map~$s$ of size~$H \times W \times D$ as follows:
\begin{equation}
    s(\mathbf{p}) = s_c  \;\;\text{for}\;\; \mathbf{p} \in \mathcal{R}_c,
\end{equation}
where we denote by~$s_c \in \mathbb{R}^{D}$ a semantic feature for a class~$c$. That is, we stack a semantic feature for a class~$c$ into corresponding regions~$\mathcal{R}_c$ labeled as the same class in the ground truth~$y$. Given the semantic feature map, the semantic encoder then outputs a semantic prototype map~$\mu$ of size~$H \times W \times C$, where
\begin{equation}
    \mu(\mathbf{p}) = \mu_c  \;\;\text{for}\;\; \mathbf{p} \in \mathcal{R}_c.
\end{equation}
We denote by~$\mu_c \in \mathbb{R}^{C}$ a semantic prototype for a class~$c$. Accordingly, we define a pixel-wise regression loss as follows:
\begin{equation} \label{eq:center}
    \mathcal{L}_\mathrm{center} = \frac{1}{\sum_{c \in \mathcal{S}} |\mathcal{R}_c|} \sum_{c \in \mathcal{S}} \sum_{\mathbf{p} \in \mathcal{R}_c} d\left( v(\mathbf{p}), \mu(\mathbf{p}) \right),
\end{equation}
where~$d(\cdot,\cdot)$ is a distance metric~(\emph{e.g.}, Euclidean distance). This term enables learning a joint embedding space by updating both encoders with a gradient of~Eq.~\eqref{eq:center}. We have observed that the semantic feature map~$s$ shows a sharp transition at object boundaries due to the stacking operation, making the semantic prototype map~$\mu$ discrete accordingly\footnote{This is because we use a $1 \times 1$ convolutional layer for the semantic encoder. Note that we could not use a CNN as the semantic encoder since it requires a ground-truth mask to obtain the semantic feature map at test time.}, as shown in~Fig.~\ref{fig:fig2}(a)~(top). By contrast, the visual feature map~$v$ smoothly varies at object boundaries due to the large receptive field of the visual encoder as shown in~Fig.~\ref{fig:fig2}(a)~(middle). That is, the visual features at object boundaries could contain a mixture of different semantics. Thus, directly minimizing~Eq.~\eqref{eq:center} might degrade performance, since this could also close the distances between semantic prototypes as shown in~Fig.~\ref{fig:fig2}(b)~(dashed lines). To address this, we exploit linearly interpolated semantic prototypes, which we refer to as virtual prototypes. The virtual prototype acts as a dustbin that gathers the visual features at object boundaries as shown in~Fig.~\ref{fig:fig2}(b)~(solid lines). However, manually interpolating semantic prototypes at all boundaries could be demanding. 

We introduce a simple yet effective implementation that gives a good compromise. Specifically, we first downsample the ground-truth mask~$y$ by a factor of~$r$ using nearest-neighbor interpolation. Similar to the previous case, we stack semantic features but with the downsampled ground-truth mask, and obtain a semantic feature map. We upsample this feature map by a factor of~$r$ again using bilinear interpolation, resulting in an interpolated one~$\tilde s$ of size~$H \times W \times D$. Given the semantic feature map~$\tilde s$, the semantic encoder outputs an interpolated semantic prototype map~$\tilde\mu$ accordingly, as shown in~Fig.~\ref{fig:fig2}(a)~(bottom). Using the interpolated semantic prototype map~$\tilde\mu$, we define a BAR loss as follows:
\begin{equation} \label{eq:bar}
    \mathcal{L}_\mathrm{bar} = \frac{1}{\sum_{c \in \mathcal{S}} |\mathcal{R}_c|} \sum_{c \in \mathcal{S}} \sum_{\mathbf{p} \in \mathcal{R}_c} d\left( v(\mathbf{p}), \tilde\mu(\mathbf{p}) \right).
\end{equation}
This term enables learning discriminative semantic prototypes. Note that it has been shown that uncertainty estimates of~\cite{UNCER20Hou} are highly activated at object boundaries. We can thus interpret the BAR loss as alleviating the influence of visual features at object boundaries in that this term encourages the visual features at object boundaries to be closer to virtual prototypes than the exact ones. Note also that~Eq.~\eqref{eq:center} is a special case of our BAR loss, that is,~$\mu = \tilde \mu$ when~$r=1$.

\begin{figure}[t]
    \centering
    \includegraphics[width=\linewidth]{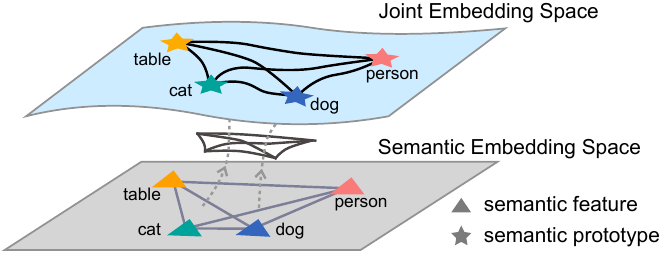}
    \vspace{-.55cm}
    \caption{\small{We visualize the relations between seen classes in semantic and joint embedding spaces. Our SC loss transfers the relations from the semantic embedding space to the joint one. This adjusts the distances between semantic prototypes explicitly, complementing the BAR loss. Best viewed in color.}}
        \vspace{-.45cm}
    \label{fig:fig3}
\end{figure}

\noindent \textbf{SC loss.}
Although CE and BAR terms help to learn discriminative representations in the joint embedding space, they do not impose explicit constraints on the distances between semantic prototypes during training. To complement this, we propose to transfer the relations of semantic features in the semantic embedding space to the semantic prototypes in the joint one. For example, we reduce the distances between semantic prototypes in the joint embedding space if corresponding semantic features are close in the semantic one~(Fig.~\ref{fig:fig3}). Concretely, we define the relation between two different classes~$i$ and~$j$ in the semantic embedding space as follows:
\begin{equation}
	r_{ij} = \frac{e^{-\tau_s d(s_{i}, s_{j})}}{\sum_{j \in \mathcal{S}} e^{-\tau_s d(s_{i}, s_{j})}},
\end{equation}
where~$\tau_s$ is a temperature parameter that controls the smoothness of relations. Similarly, we define the relation in the joint embedding space as follows:
\begin{equation}
	\hat r_{ij} = \frac{e^{- \tau_\mu d(\mu_{i}, \mu_{j})}}{\sum_{j \in \mathcal{S}} e^{- \tau_\mu d(\mu_{i}, \mu_{j})}},
\end{equation}
where~$\tau_\mu$ is a temperature parameter. To encourage the consistency between two embedding spaces, we define a SC loss as follows:
\begin{equation}
    \mathcal{L}_\mathrm{sc} = -\sum_{i \in \mathcal{S}} \sum_{j \in \mathcal{S}} r_{ij} \log \frac{\hat r_{ij}}{r_{ij}}.
\end{equation}
This term regularizes the distances between semantic prototypes of seen classes. Similarly, CSRL~\cite{CSRL20Li} distills the relations of real visual features to the synthesized ones. It however exploits semantic features of unseen classes during training, suggesting that both generator and classifier should be trained again to handle novel unseen classes.

\subsection{Inference}\vspace{-.1cm} \label{sec:sec3_3}
\begin{figure}[t]
	\centering
    \begin{subfigure}{0.49\linewidth}
        \centering
        \includegraphics[width=\linewidth]{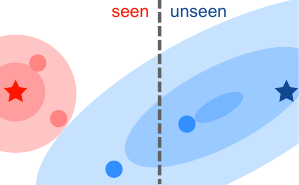}
	\end{subfigure}
    \begin{subfigure}{0.49\linewidth}
        \centering
		\includegraphics[width=\linewidth]{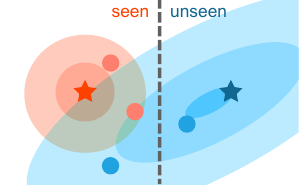}
    \end{subfigure}
    \begin{subfigure}{\linewidth}
        \centering
        \caption{\small{Visualization of a seen bias problem.}}
    \end{subfigure}
    \begin{subfigure}{0.49\linewidth}
        \centering
        \includegraphics[width=\linewidth]{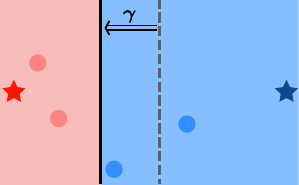}
	\end{subfigure}
    \begin{subfigure}{0.49\linewidth}
        \centering
		\includegraphics[width=\linewidth]{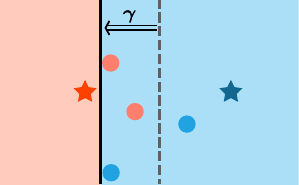}
    \end{subfigure}
    \begin{subfigure}{\linewidth}
        \centering
        \caption{\small{CS~\cite{CALIB16Chaqo}.}}
    \end{subfigure}
    \begin{subfigure}{0.49\linewidth}
        \centering
        \includegraphics[width=\linewidth]{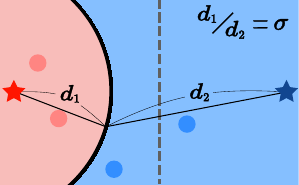}
    \end{subfigure}
    \begin{subfigure}{0.49\linewidth}
        \centering
        \includegraphics[width=\linewidth]{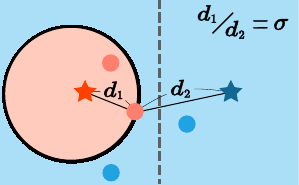}
	\end{subfigure}
	\begin{subfigure}{\linewidth}
        \centering
        \caption{\small{AC.}}
    \end{subfigure}
	\vspace{-.65cm}
    \caption{\small{Comparison of CS~\cite{CALIB16Chaqo} and AC. We visualize semantic prototypes and visual features by stars and circles, respectively. The decision boundary of the NN classifier is shown as dashed lines.~(a) We show the seen bias problem with the distribution of visual features in two cases. One is when two different semantic prototypes are distant~(left), and the other is the opposite situation~(right). Note that visual features of seen classes are tightly clustered, while those of unseen classes are skewed.~(b) CS shifts the decision boundary to semantic prototypes of seen classes. Although CS alleviates the seen bias problem~(left), it might degrade performance for seen classes~(right). Thus, the value of~$\gamma$ should be chosen carefully.~(c) We modulate the decision boundary with the Apollonius circle. This gives a good compromise between improving performance for unseen classes~(left) and preserving that for seen ones~(right). Best viewed in color.}}
    \vspace{-.45cm}
    \label{fig:fig4}
\end{figure}

Our discriminative approach enables handling semantic features of arbitrary classes at test time without re-training, which is suitable for real-world scenarios. Specifically, the semantic encoder takes semantic features of both seen and unseen classes, and outputs corresponding semantic prototypes. We then compute the distances from individual visual features to each semantic prototype. That is, we formulate the inference process as a retrieval task using the semantic prototypes as a NN classifier in the joint embedding space. A straightforward way to classify each visual feature\footnote{We upsample~$v$ into the image resolution~$H_o \times W_o$ using bilinear interpolation for inference.} is to assign the class of its nearest semantic prototype as follows:
\begin{equation} \label{eq:1stnn}
    \hat{y}_{\mathrm{nn}}(\mathbf{p}) = \operatorname*{argmin}_{c \in \mathcal{S} \cup \mathcal{U}} d(v(\mathbf{p}), \mu_{c}).
\end{equation}
Although our approach learns discriminative visual features and semantic prototypes, visual features of unseen classes might still be biased towards those of seen classes~(Fig.~\ref{fig:fig4}(a)), especially when both have similar appearance. For example, a cat (a unseen object class) is more likely to be predicted as a dog (a seen one). To address this, the work of~\cite{CALIB16Chaqo} proposes a calibrated stacking~(CS) method that penalizes scores of seen classes with a constant value. In our case, this can be formulated with an adjustable parameter~$\gamma$ as follows:
\begin{equation} \label{eq:calib}
    \hat{y}_{\mathrm{cs}}(\mathbf{p}) = \operatorname*{argmin}_{c \in \mathcal{S} \cup \mathcal{U}}d(v(\mathbf{p}), \mu_{c}) - \gamma \boldsymbol{\mathbbm{1}}[c \in \mathcal{U}],
\end{equation}
where we denote by~$\boldsymbol{\mathbbm{1}}[\cdot]$ an indicator function whose value is~$1$ if the argument is true, and~$0$ otherwise. We interpret this as shifting the decision boundary of the NN classifier to semantic prototypes of seen classes. CS alleviates the seen bias problem when the first two nearest prototypes of a particular visual feature are distant as shown in~Fig.~\ref{fig:fig4}(b)~(left). It however applies the same value of~$\gamma$ to the case when the first two nearest prototypes are close as shown in~Fig.~\ref{fig:fig4}(b)~(right), degrading performance for seen classes. Finding the best value of~$\gamma$ is thus not trivial. Instead of shifting, we propose to modulate the decision boundary using the Apollonius circle. Specifically, we first compute the distances to the first two nearest semantic prototypes for individual visual features as follows: 
\begin{equation}
	d_1(\mathbf{p}) = d(v(\mathbf{p}), \mu_{c_\mathrm{1st}})
	\;\,\text{and}\,\; 
	d_2(\mathbf{p}) = d(v(\mathbf{p}), \mu_{c_\mathrm{2nd}}),
\end{equation}
where~$0 < d_1(\mathbf{p}) \leq d_2(\mathbf{p})$. We denote by~$c_\mathrm{1st}$ and~$c_\mathrm{2nd}$ the class of the first and second nearest prototype, respectively. We then define the Apollonius circle, which is used as our decision boundary, with an adjustable parameter~$\sigma$ as follows:
\begin{equation}
	\mathcal{A}(\sigma) =\{\mathbf{p} \mid d_1(\mathbf{p}):d_2(\mathbf{p}) = \sigma : 1 \},
\end{equation}
where we denote by~$\mathcal{A}(\sigma)$ the boundary of the Apollonius circle. The decision rule is defined with this circle as follows:
\begin{equation} \label{eq:ac}
	\hat{y}_{\mathrm{ac}}(\mathbf{p}) = 
	\left\{\begin{aligned} 
	c_\mathrm{12}(\mathbf{p}) 
	&\;,\; c_\mathrm{1st} \in \mathcal{S} \;\text{and}\; c_\mathrm{2nd} \in \mathcal{U}
	\\
	c_\mathrm{1st} 
	&\;,\; \text{otherwise}
	\end{aligned}\right.,
\end{equation}
where
\begin{equation}
	c_\mathrm{12}(\mathbf{p}) = 
	c_\mathrm{1st}\boldsymbol{\mathbbm{1}}\left[\frac{d_1(\mathbf{p})}{d_2(\mathbf{p})} \leq \sigma\right] + 
	c_\mathrm{2nd}\boldsymbol{\mathbbm{1}}\left[\frac{d_1(\mathbf{p})}{d_2(\mathbf{p})} > \sigma\right].
\end{equation}
\begin{table*}[t]
    \centering
    \footnotesize
    \caption{\small{Quantitative results on the PASCAL VOC~\cite{VOC10Ever} and Context~\cite{CONTEXT14Motta} validation sets in terms of mIoU. Numbers in bold are the best performance and underlined ones are the second best. We report our average scores over five runs with standard deviations in parentheses.}}
	\vspace{-.2cm}
    \label{table:sota}
    \addtolength{\tabcolsep}{-6.pt}
    \begin{tabular}{p{1cm}|p{1.5cm}|p{1.cm}p{1.cm}p{1.cm}|p{1.cm}p{1.cm}p{1.cm}|p{1.cm}p{1.cm}p{1.cm}|p{1.cm}p{1.cm}p{1.cm}|p{1.cm}p{1.cm}p{1.cm}}
    \hlineB{2.5}
    \hfil\multirow{2}{*}{Datasets} & \multirow{2}{*}{Methods} & \multicolumn{3}{c|}{unseen-2} & \multicolumn{3}{c|}{unseen-4} & \multicolumn{3}{c|}{unseen-6} & \multicolumn{3}{c|}{unseen-8} & \multicolumn{3}{c}{unseen-10} \\ \cline{3-17}
    && \hfil $\text{mIoU}_\mathcal{S}$ & \hfil $\text{mIoU}_\mathcal{U}$ & \hfil \text{hIoU} & \hfil $\text{mIoU}_\mathcal{S}$ & \hfil $\text{mIoU}_\mathcal{U}$ & \hfil \text{hIoU} & \hfil $\text{mIoU}_\mathcal{S}$ & \hfil $\text{mIoU}_\mathcal{U}$ & \hfil \text{hIoU} & \hfil $\text{mIoU}_\mathcal{S}$ & \hfil $\text{mIoU}_\mathcal{U}$ & \hfil \text{hIoU} & \hfil $\text{mIoU}_\mathcal{S}$ & \hfil $\text{mIoU}_\mathcal{U}$ & \hfil \text{hIoU}\\
    \hline
    \hfil\multirow{5}{*}{VOC} & DeViSE~\cite{DEVISE13Frome} & \hfil 68.1 & \hfil 3.2  & \hfil 6.1  & \hfil 64.3 & \hfil 2.9  & \hfil 5.5  & \hfil 39.8 & \hfil 2.7  & \hfil 5.1  & \hfil 35.7 & \hfil 2.0  & \hfil 3.8  & \hfil 31.7 & \hfil 1.9  & \hfil 3.6 \\
    & SPNet~\cite{SPNET19Xian}  & \hfil 71.8 & \hfil 34.7 & \hfil 46.8 & \hfil 67.3 & \hfil 21.8 & \hfil 32.9 & \hfil 64.5 & \hfil 20.1 & \hfil 30.6 & \hfil 61.2 & \hfil 19.9 & \hfil 30.0 & \hfil 59.0 & \hfil 18.1 & \hfil 27.7 \\ 
    & ZS3Net~\cite{ZS3NET19Bucher} & \hfil 72.0 & \hfil 35.4 & \hfil 47.5 & \hfil 66.4 & \hfil 23.2 & \hfil 34.4 & \hfil 47.3 & \hfil 24.2 & \hfil 32.0 & \hfil 29.2 & \hfil 22.9 & \hfil 25.7 & \hfil 33.9 & \hfil 18.1 & \hfil 23.6 \\ 
    & CSRL~\cite{CSRL20Li}   & \hfil 73.4 & \hfil 45.7 & \hfil \textbf{56.3} & \hfil 69.8 & \hfil 31.7 & \hfil \underline{43.6} & \hfil 66.2 & \hfil 29.4 & \hfil \underline{40.7} & \hfil 62.4 & \hfil 26.9 & \hfil \underline{37.6} & \hfil 59.2 & \hfil 21.0 & \hfil \underline{31.0} \\
    & Ours & \hfil 68.9 \tiny{(1.0)} & \hfil 43.2 \tiny{(0.9)} & \hfil \underline{53.1} \tiny{(0.4)} & \hfil 67.0 \tiny{(1.2)} & \hfil 33.4 \tiny{(0.4)} & \hfil \textbf{44.6} \tiny{(0.3)} & \hfil 63.2 \tiny{(0.4)} & \hfil 30.5 \tiny{(0.3)} & \hfil \textbf{41.1} \tiny{(0.2)} & \hfil 58.5 \tiny{(0.9)} & \hfil 29.0 \tiny{(0.8)} & \hfil \textbf{38.8} \tiny{(0.6)} & \hfil 63.5 \tiny{(0.4)} & \hfil 22.5 \tiny{(0.4)} & \hfil \textbf{33.2} \tiny{(0.4)} 
    \\
    \hline\hline
    \hfil\multirow{5}{*}{Context} & DeViSE~\cite{DEVISE13Frome} & \hfil 35.8 & \hfil 2.7  & \hfil 5.0  & \hfil 33.4 & \hfil 2.5  & \hfil 4.7  & \hfil 31.9 & \hfil 2.1  & \hfil 3.9  & \hfil 22.0 & \hfil 1.7  & \hfil 3.2  & \hfil 17.5 & \hfil 1.3  & \hfil 2.4 \\
    & SPNet~\cite{SPNET19Xian}  & \hfil 38.2 & \hfil 16.7 & \hfil 23.2 & \hfil 36.3 & \hfil 18.1 & \hfil 24.2 & \hfil 31.9 & \hfil 19.9 & \hfil 24.5 & \hfil 28.6  & \hfil 14.3 & \hfil 19.1 & \hfil 27.1 & \hfil 9.8  & \hfil 14.4 \\ 
    & ZS3Net~\cite{ZS3NET19Bucher} & \hfil 41.6 & \hfil 21.6 & \hfil 28.4 & \hfil 37.2 & \hfil 24.9 & \hfil 29.8 & \hfil 32.1 & \hfil 20.7 & \hfil 25.2 & \hfil 20.9 & \hfil 16.0 & \hfil 18.1 & \hfil 20.8 & \hfil 12.7 & \hfil 15.8 \\ 
    & CSRL~\cite{CSRL20Li}   & \hfil 41.9 & \hfil 27.8 & \hfil \underline{33.4} & \hfil 39.8 & \hfil 23.9 & \hfil \underline{29.9} & \hfil 35.5 & \hfil 22.0 & \hfil \underline{27.2} & \hfil 31.7 & \hfil 18.1 & \hfil \underline{23.0} & \hfil 29.4 & \hfil 14.6 & \hfil \underline{19.5} \\
    & Ours & \hfil 38.2 \tiny{(1.2)} & \hfil 32.9 \tiny{(1.4)} & \hfil \textbf{35.3} \tiny{(0.9)} & \hfil 36.9 \tiny{(0.8)} & \hfil 30.7 \tiny{(1.5)} & \hfil \textbf{33.5} \tiny{(0.7)} & \hfil 36.2 \tiny{(0.6)} & \hfil 23.2 \tiny{(0.4)} & \hfil \textbf{28.3} \tiny{(0.4)} & \hfil 32.4 \tiny{(0.9)} & \hfil 20.2 \tiny{(0.4)} & \hfil \textbf{24.9} \tiny{(0.3)} & \hfil 33.0 \tiny{(0.6)} & \hfil 14.9 \tiny{(0.7)} & \hfil \textbf{20.5} \tiny{(0.6)} 
    \\
    \hlineB{2.5}
    \end{tabular}
    \vspace{-.45cm}
\end{table*}That is, we assign~$c_\mathrm{1st}$ and~$c_\mathrm{2nd}$ to the visual features inside and outside the Apollonius circle, respectively, providing a better compromise between performance for seen and unseen classes. This is more intuitive than CS in that visual features of seen classes are tightly centered around corresponding semantic prototypes, while those of unseen classes are distorted and dispersed~(Fig.~\ref{fig:fig4}(a)). Note that the radius of this circle adaptively changes in accordance with the distance between the first two nearest semantic prototypes\footnote{Please refer to the supplementary material for a more detailed description of this.}. As shown in Fig.~\ref{fig:fig4}(c), this enables reducing the seen bias problem in both cases, while maintaining performance for seen classes even with the same value of~$\sigma$~(right). Furthermore, unlike CS, we modulate the decision boundary, only when the class of the first and second nearest semantic prototype belongs to~$\mathcal{S}$ and~$\mathcal{U}$, respectively, since the seen bias problem is most likely to occur in this case. Note that AC reduces to the NN classifier in~Eq.~\eqref{eq:1stnn} when the adjustable parameter~$\sigma=1$.

\section{Experiments}
\vspace{-.1cm}
\subsection{Implementation details}
\vspace{-.1cm}\noindent \textbf{Dataset and evaluation.}
We perform experiments on standard GZS3 benchmarks: PASCAL VOC~\cite{VOC10Ever} and PASCAL Context~\cite{CONTEXT14Motta}. The PASCAL VOC dataset provides~$1,464$ training and $1,449$ validation samples of~$20$ object classes, while the PASCAL Context dataset contains~$4,998$ training and~$5,105$ validation samples of~$59$ thing and stuff classes. Both datasets include a single background class, resulting in~$21$ and~$60$ classes in total, respectively. Following the common practice in~\cite{ZS3NET19Bucher,CAGNET20Gu,CSRL20Li,SPNET19Xian}, we use augmented~$10,582$ training samples~\cite{AUG11Hari} for PASCAL VOC. We follow the experiment settings provided by~ZS3Net~\cite{ZS3NET19Bucher}. It provides five splits for each dataset, where each split contains previous unseen classes gradually as follows: (1) 2-cow/motorbike, (2) 4-airplane/sofa, (3) 6-cat/tv, (4) 8-train/bottle, (5) 10-chair/potted-plant for PASCAL VOC, and (1) 2-cow/motorbike, (2) 4-sofa/cat, (3) 6-boat/fence, (4) 8-bird/tvmonitor, (5) 10-keyborad/aeroplane for PASCAL Context. In all experiments, we exclude training samples that contain unseen classes, and adopt~\textit{word2vec}~\cite{W2V13Mikolov} obtained from the names of corresponding classes as semantic features, whose dimension is 300. For evaluation, we use the mean intersection-over-union~(mIoU) metric. In detail, we provide mIoU scores for sets of seen and unseen classes, denote by~$\text{mIoU}_\mathcal{S}$ and~$\text{mIoU}_\mathcal{U}$, respectively. Since the arithmetic mean might be dominated by~$\text{mIoU}_\mathcal{S}$, we compute the harmonic mean (hIoU) of~$\text{mIoU}_\mathcal{S}$ and~$\text{mIoU}_\mathcal{U}$.
We do not apply dCRF~\cite{CRF12Kr} and a test-time augmentation strategy during inference. Note that we present more results including the experiment settings provided by~SPNet~\cite{SPNET19Xian} in the supplementary material.

\vspace{-.21cm}\noindent \textbf{Training.}
For fair comparison, we use DeepLabV3+~\cite{V3PLUS18Chen} with ResNet-101~\cite{RES16He} as our visual encoder. Following ZS3Net~\cite{ZS3NET19Bucher}, ResNet-101 is initialized with the pre-trained weights for ImageNet classification~\cite{IMAGENET09Deng}, where training samples of seen classes are used only. We train the visual encoder using the SGD optimizer with learning rate, weight decay, and momentum of 2.5e-4, 1e-4, and 0.9, respectively. We adopt a linear layer as the semantic encoder, and train it using the Adam optimizer with learning rate of 2e-4. The entire model is trained for 50 and 200 epochs with a batch size of 32 on PASCAL VOC~\cite{VOC10Ever} and Context~\cite{CONTEXT14Motta}, respectively. We use the poly schedule to adjust the learning rate. In all experiments, we adopt a Euclidean distance for~$d(\cdot,\cdot)$. 

\vspace{-.21cm}\noindent \textbf{Hyperarameters.}
We empirically set~$(r,\tau_s, \tau_\mu)$ to~$(4,5,1)$ and~$(4,7,1)$ for PASCAL VOC~\cite{VOC10Ever} and Context~\cite{CONTEXT14Motta}, respectively. Other parameters~($\lambda, \sigma$) are chosen by cross-validation for each split as in~\cite{CROSSVALID117Bucher}. We provide a detailed analysis on these parameters in the supplementary material.

\subsection{Results}
\vspace{-.1cm}We compare in Table~\ref{table:sota} our approach with state-of-the-art GZS3 methods on PASCAL VOC~\cite{VOC10Ever} and Context~\cite{CONTEXT14Motta}. We report average scores over five runs with standard deviations. All numbers for other methods are taken from~CSRL~\cite{CSRL20Li}. From this table, we have three findings as follows: (1) Our approach outperforms SPNet~\cite{SPNET19Xian} on both datasets by a considerable margin in terms of~$\text{mIoU}_\mathcal{U}$ and hIoU. This confirms that exploiting a joint embedding space enables learning better representations. (2) We achieve a new state of the art on four out of five PASCAL VOC splits. Although CSRL shows better results on the unseen-2 split, they require semantic features of unseen classes during training. This suggests that both generator and classifier of CSRL should be retrained whenever novel unseen classes appear, which is time consuming. Our discriminative approach is more practical in that the semantic encoder takes semantic features of arbitrary classes without the retraining process. (3) We can clearly see that our approach outperforms all other methods including the generative methods~\cite{ZS3NET19Bucher,CSRL20Li} on all splits of PASCAL Context. A plausible reason is that PASCAL Context contains four times more seen classes including stuff ones than VOC. This makes the generative methods suffer from a severe bias problem towards seen classes. 

\subsection{Discussion}\vspace{-.1cm}
\noindent \textbf{Ablation study.}
In the first four rows of Table~\ref{table:ablation}, we present an ablation analysis on different losses in our framework. We adopt a simple NN classifier to focus on the effect of each term. Since the CE loss is crucial to learn discriminative visual features, we incorporate it to all variants. To the baseline, we report mIoU scores without both~$\mathcal{L}_\mathrm{bar}$ and~$\mathcal{L}_\mathrm{sc}$,~\emph{i.e.},~$r=1$ and~$\lambda=0$, in the first row. The second row shows that the BAR loss gives a hIoU gain of 0.9\% over the baseline. This is significant in that the difference between the first two rows is whether a series of two interpolations is applied to a semantic feature map or not, before inputting it to a semantic encoder~(see Sec.~\ref{sec:sec3_2}). We can also see that explicitly regularizing the distances between semantic prototypes improves performance for unseen classes in the third row. The fourth row demonstrates that BAR and SC terms are complementary to each other, achieving the best performance.

\begin{figure}[t]	
	\centering
    \begin{subfigure}{.49\linewidth}
        \centering
        \includegraphics[width=\linewidth]{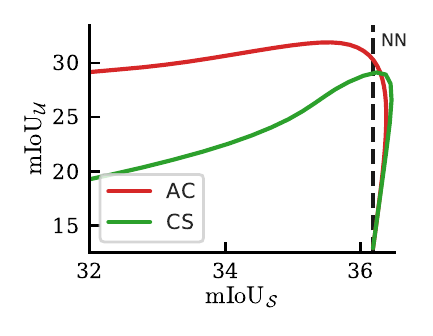}
	\end{subfigure}
    \begin{subfigure}{.49\linewidth}
        \centering
		\includegraphics[width=\linewidth]{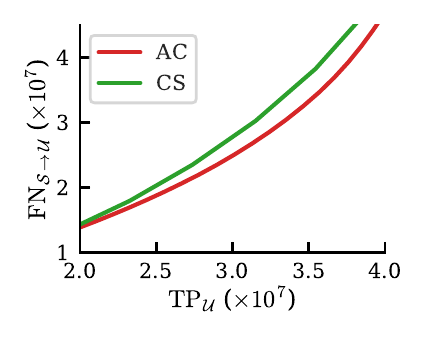}
    \end{subfigure}
	\vspace{-.5cm}
    \caption{\small{Comparison of CS~\cite{CALIB16Chaqo} and AC by varying~$\gamma$ and~$\sigma$, respectively, on the unseen-4 split of PASCAL Context~\cite{CONTEXT14Motta}. We show the mIoU$_\mathcal{U}$-mIoU$_\mathcal{S}$ curves~(left), and how FN$_{\mathcal{S}\rightarrow\mathcal{U}}$ changes w.r.t. TP$_\mathcal{U}$~(right). Best viewed in color.}}
    \label{fig:fig5}
    \vspace{-.15cm}
\end{figure}

\noindent \textbf{Comparison with CS.}
The last two rows in Table~\ref{table:ablation} show a quantitative comparison of CS~\cite{CALIB16Chaqo} and AC in terms of mIoU scores. We can see that both CS and AC improve performance for unseen classes by large margins. A reason is that visual features for unseen classes are skewed and biased towards those of seen classes~(Fig.~\ref{fig:fig4}(a)). It is worth noting that AC further achieves a mIoU$_\mathcal{U}$ gain of 2.7\% over CS with a negligible overhead, demonstrating the effectiveness of using the Apollonius circle. In Fig.~\ref{fig:fig5}, we plot performance variations according to the adjustable parameter for each method,~\emph{i.e.},~$\gamma$ and~$\sigma$, in the range of~$[0,12]$ and~$(0,1]$ with intervals of~$0.5$ and~$0.05$ for CS and AC, respectively. We first compare the mIoU$_\mathcal{U}$-mIoU$_\mathcal{S}$ curves in Fig.~\ref{fig:fig5} (left). For comparison, we visualize the mIoU$_\mathcal{S}$ of the NN classifier by a dashed line. We can see that AC always gives better mIoU$_\mathcal{U}$ scores for all mIoU$_\mathcal{S}$ values on the left-hand side of the dashed line, suggesting that AC is more robust w.r.t. the adjustable parameter. We also show that how false negatives of seen classes change according to true positives of unseen classes (TP$_\mathcal{U}$) in Fig.~\ref{fig:fig5}~(right). In particular, we compute false negatives of seen classes, when they are predicted as one of unseen classes, denoted by FN$_{\mathcal{S}\rightarrow\mathcal{U}}$. We can clearly see that CS has more FN$_{\mathcal{S}\rightarrow\mathcal{U}}$ than AC at the same value of TP$_\mathcal{U}$, confirming once again that AC is more robust to the parameter, while providing better results.

\noindent \textbf{Analysis of embedding spaces.}
To verify that exploiting a joint embedding space alleviates a seen bias problem, we compare in Table~\ref{table:joint} variants of our approach with ZS3Net~\cite{ZS3NET19Bucher}. First, we attempt to project visual features to corresponding semantic ones without exploiting a semantic encoder. This, however, provides a trivial solution that all visual features are predicted as a background class. Second, we adopt a two-stage discriminative approach, that is, training visual and semantic encoders sequentially. We first train a segmentation network that consists of a feature extractor and a classifier with seen classes. The learned feature extractor is then fixed and it is used as a visual encoder to train a semantic encoder~(`S$\rightarrow$V'). We can see from the first two rows that this simple variant with BAR and SC terms already outperforms ZS3Net, demonstrating the effectiveness of the discriminative approach. These variants are, however, outperformed by our approach that gives the best hIoU score of~$44.6$~(Table~\ref{table:sota}). To further verify our claim, we train the generator of ZS3Net using visual features extracted from our visual encoder~(`ZS3Net$^\ddagger$'). For comparison, we also report the results obtained by our implementation of ZS3Net~(`ZS3Net$^\dagger$'). From the last two rows, we can clearly see that `ZS3Net$^\ddagger$' outperforms `ZS3Net$^\dagger$'. This confirms that our approach alleviates the seen bias problem, enhancing the generalization ability of visual features.

\begin{table}[t]
    \centering
    \footnotesize
    \caption{\small{Comparison of mIoU scores using different loss terms and inference techniques on the unseen-4 split of PASCAL Context~\cite{CONTEXT14Motta}. For an ablation study on different loss terms, we use a NN classifier without applying any inference techniques in order to focus more on the effect of each term. CS: calibrated stacking~\cite{CALIB16Chaqo}; AC: Apollonius circle.}}
	\vspace{-.2cm}
    \label{table:ablation}
    \addtolength{\tabcolsep}{-4.0pt}
    \begin{tabular}{c|cc|c|cc|p{1.1cm}p{1.1cm}p{1.1cm}}
    \hlineB{2.5}
    $\mathcal{L}_\mathrm{ce}$ & $\mathcal{L}_\mathrm{center}$ & $\mathcal{L}_\mathrm{bar}$ & $\mathcal{L}_\mathrm{sc}$ & $\text{CS}$ & $\text{AC}$ & \hfil $\text{mIoU}_\mathcal{S}$ & \hfil $\text{mIoU}_\mathcal{U}$ & \hfil \text{hIoU} \\
    \hline
    \checkmark & \checkmark &            &            & & & \hfil 37.7 & \hfil 10.0 & \hfil 15.8 \\
    \checkmark &            & \checkmark &            & & & \hfil 37.9 & \hfil 10.7 & \hfil 16.7 \\
    \checkmark & \checkmark &            & \checkmark & & & \hfil 36.1 & \hfil 11.8 & \hfil 17.8 \\
    \checkmark &            & \checkmark & \checkmark & & & \hfil 36.2 & \hfil 12.9 & \hfil 19.0 \\
    \hline
    \checkmark &            & \checkmark & \checkmark & \checkmark & & \hfil 36.2 & \hfil 29.1 & \hfil \underline{32.3} \\
    \checkmark &            & \checkmark & \checkmark & & \checkmark & \hfil 35.7 & \hfil 31.8 & \hfil \textbf{33.7} \\
    \hlineB{2.5}
    \end{tabular}
    \vspace{-.15cm}
\end{table}

\begin{table}[t]
    \centering
    \footnotesize
    \caption{\small{Quantitative comparison on the unseen-4 split of PASCAL VOC~\cite{VOC10Ever}.~$\dagger$: reimplementation;~$\ddagger$: our visual encoder.}}
	\vspace{-.2cm}
    \label{table:joint}
    \addtolength{\tabcolsep}{-1.0pt}
    \begin{tabular}{p{2.3cm}|p{1.3cm}p{1.3cm}p{1.3cm}}
    \hlineB{2.5}
    Methods & \hfil $\text{mIoU}_\mathcal{S}$ & \hfil $\text{mIoU}_\mathcal{U}$ & \hfil $\text{hIoU}$
     \\
    \hline
    S$\rightarrow$V: $\mathcal{L}_{\mathrm{center}}$& \hfil 61.7 & \hfil 20.9 & \hfil 31.2
    \\
    S$\rightarrow$V: $\mathcal{L}_{\mathrm{bar}}$ + $\mathcal{L}_{\mathrm{sc}}$ & \hfil 65.7 & \hfil 30.3 & \hfil \underline{41.5}
    \\
    \hline
    ZS3Net~\cite{ZS3NET19Bucher} & \hfil 66.4 & \hfil 23.2 & \hfil 34.4 
    \\
    ZS3Net$^\dagger$             & \hfil 68.8 & \hfil 28.8 & \hfil \underline{40.6}
    \\
    ZS3Net$^\ddagger$            & \hfil 68.5 & \hfil 31.8 & \hfil \textbf{43.4}
	\\
    \hlineB{2.5}
    \end{tabular}
	\vspace{-.45cm}
\end{table}

\section{Conclusion}
We have introduced a discriminative approach, dubbed JoEm, that overcomes the limitations of generative ones in a unified framework. We have proposed two complementary losses to better learn representations in a joint embedding space. We have also presented a novel inference technique using the circle of Apollonius that alleviates a seen bias problem significantly. Finally, we have shown that our approach achieves a new state of the art on standard GZS3 benchmarks.

\noindent \textbf{Acknowledgments.}
This work was supported in part by the National Research Foundation of Korea (NRF) grant funded by the Korea government (MSIP) (NRF-2019R1A2C2084816) and the Yonsei University Research Fund of 2021 (2021-22-0001).

{\small
\bibliographystyle{ieee_fullname}
\bibliography{egbib}
}

\clearpage


\section*{Supplementary material}

# Exploiting a Joint Embedding Space for Generalized Zero-Shot Semantic Segmentation
## Supplement

Donghyeon Baek[*]  Youngmin Oh[*]  Bumsub Ham[†]
School of Electrical and Electronic Engineering, Yonsei University
https://cvlab.yonsei.ac.kr/projects/JoEm

Here we present a detailed description of the Apollonius circle in Sec. 1, and show more results on different experiment settings in Sec. 2.

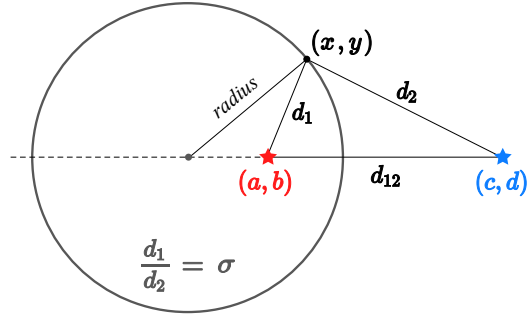

Figure 1. Illustration of the Apollonius circle in a two-dimensional Euclidean space.

## 1. Apollonius' definition of a circle

A circle in a Euclidean space refers to a set of points, where all distances from each point to a particular point (*i.e.*, *center*) are equal. Apollonius of Perga, on the other hand, defines a circle as a set of points that have the same distance ratio for two particular points (Fig. 1). Concretely, the Apollonius circle in a two-dimensional Euclidean space is defined as follows:

$$\begin{aligned}
&\frac{d_1}{d_2} = \sigma \\
&\iff \frac{\sqrt{(x-a)^2 + (y-b)^2}}{\sqrt{(x-c)^2 + (y-d)^2}} = \sigma \\
&\iff (x-a)^2 + (y-b)^2 = \sigma^2 \left((x-c)^2 + (y-d)^2\right) \\
&\iff \left(x - \frac{a - \sigma^2 c}{1 - \sigma^2}\right)^2 + \left(y - \frac{b - \sigma^2 d}{1 - \sigma^2}\right)^2 = \frac{\sigma^2}{(1-\sigma^2)^2}\left((a-c)^2 + (b-d)^2\right),
\end{aligned} \quad (1)$$

where $(a, b)$ and $(c, d)$ are the two particular points, often called *foci*. From the last row in Eq. (1), we reformulate the radius of this circle as follows:

$$\begin{aligned}
radius &= \frac{\sigma}{1-\sigma^2}\sqrt{(a-c)^2 + (b-d)^2} \\
&= \frac{\sigma}{1-\sigma^2} d_{12}
\end{aligned} \quad (2)$$

We can see that the radius is proportional to the distance between the *foci*, *i.e.*, $d_{12}$. This confirms that Apollonius calibration handles the seen bias problem adaptively, even with the same value of $\sigma$ (see Figure 4 in the main paper).

---

[*]Equal contribution, [†]Corresponding author.

Table 1. Comparison of different experiment settings on PASCAL VOC [6]. Exclude(*classes*): exclude all samples that contain at least one of the *classes*; Ignore(*classes*): mark regions of the *classes* in ground-truth masks with *void* labels; IN: pre-trained weights for ImageNet classification [5]; IN*: pre-trained weights for ImageNet classification only with seen classes.

| Settings | # of splits | Training | Inference | Architecture (backbone) | Initialization | Semantic features |
|---|---|---|---|---|---|---|
| ZS3Net [2] | 5 | Exclude(unseen classes) | - | DeepLabV3+ [4] (ResNet-101 [8]) | IN* | *word2vec* [10] |
| SPNet [12] | 1 | Ignore(background & unseen classes) | Ignore(background class) | DeepLabV2 [3] (ResNet-101 [8]) | IN | *word2vec* [10] & *fastText* [9] |

## 2. More results

Table 1 summarizes differences between experiment settings provided by ZS3Net [2] and SPNet [12] on PASCAL VOC [6]. The main differences between them lie in training and inference processes. ZS3Net excludes training samples that contain at least one of the unseen classes. This, however, would leave a small number of training samples only, since objects co-occur frequently. For example, almost half of the training samples are eliminated for the unseen-10 split $(10,582 \rightarrow 5,408)$. To take into account this problem, SPNet uses all training samples but ignores pixels of unseen classes, which is more feasible for the task of semantic segmentation. Note that SPNet however ignores the background class during both training and inference. That is, SPNet requires pixel-wise annotations for the background class to discriminate foreground objects from a background during inference, while ZS3Net does not impose any assumptions for inference. This is why we have followed the setting of ZS3Net in the main paper. In the following, we present qualitative and quantitative results for each experiment setting.

### 2.1. ZS3Net.

We vary the value of $r$ to analyze its effect on the unseen-4 split of PASCAL VOC [6] and Context [11] in Fig. 2. We can see that using $r > 1$ always gives better results than $r = 1$, confirming again the superiority of our BAR loss. We empirically set $r$ to 4 in all experimental settings provided by ZS3Net [2] in order to generate more virtual prototypes. Since the spatial size of feature maps obtained from DeepLabV3+ is already small (*e.g.*, $78 \times 78$), we do not use higher values of $r$, *i.e.*, $r > 4$, to maintain the spatial size. For temperatures in Eqs. (7) and (8), we first fix $\tau_s$ as one in order to reduce the cost of selecting these hyperparmeters. Then, we find a value of $\tau_\mu$ that allows the highest relation probability in Eq. (8) to be around 0.9. As a result, we empirically set $\tau_\mu$ to 5 and 7 for all experiments on PASCAL VOC and Context, respectively. We also analyze effects of other hyperparameters $(\lambda, \sigma)$ on PASCAL VOC and Context in Figs. 3 and 4, respectively. We first use a grid search to set $\lambda \in \{10^{-3}, 10^{-2}, 10^{-1}, 1, 10\}$ with fixing the adjustable parameter $\sigma$ to 0.5. From the top rows of Figs. 3 and 4, we can clearly see that our approach is robust to changes of the balance parameter $\lambda$. We then vary $\sigma$ in the range of $(0, 1)$ with an interval of 0.1. From the bottom rows of Figs. 3 and 4, we can also see that Apollonius calibration brings considerable performance improvement in the range of $[0.5, 1)$. This method, however, degrades performance in the range of $(0, 0.5)$. A plausible reason is that the radius of the Apollonius circle becomes small as in Eq. (2), increasing classification errors. It is worth noting that the values of hyperparameter used in the main paper (dotted lines) are not optimal, since we have adopted a cross-validation [1] for each split.

We compare in Table 2 per-class mIoU scores on the unseen-4 split of PASCAL VOC. We can see that ZS3Net [2] obtains mIoU$_\mathcal{U}$ scores at the cost of mIoU$_\mathcal{S}$ ones. For example, the mIoU score of a sheep class is 0. We reimplement ZS3Net ('ZS3Net[†]'), and find that ZS3Net requires a scaling factor for unseen classes to compute the CE loss during retraining. This scaling factor is particularly important for improving performance in that ZS3Net[†] outperforms ZS3Net by simply changing the value of the scaling factor from 100 to 10. We also report per-class results of ZS3Net using our visual encoder ('ZS3Net[‡]'). This model shows higher mIoU$_\mathcal{U}$ scores than ZS3Net[†] for cow, motorbike, and sofa classes, demonstrating once again that our approach alleviates the seen bias problem (see Table 3 in the main paper).

We provide in Fig. 5 visual examples using different losses in our framework. To the baseline, we show results without BAR and SC losses in the third column. We can see that BAR and SC losses give better results than the baseline in the fourth and fifth columns, respectively. In the two rightmost columns, we show BAR and SC losses complement each other and Apollonius calibration considerably reduces the seen bias problem, respectively. In Figs. 7 and 8, we present qualitative results on the unseen-4 split of PASCAL VOC and Context, respectively. We can see that our approach provides more accurate results than ZS3Net, especially for unseen classes.

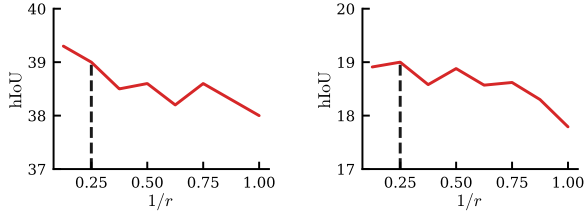

Figure 2. Comparison of hIoU scores for different values of $r$ on the unseen-4 split of PASCAL VOC [6] (left) and PASCAL Context [11] (right). The dotted line indicates the value used in the main paper.

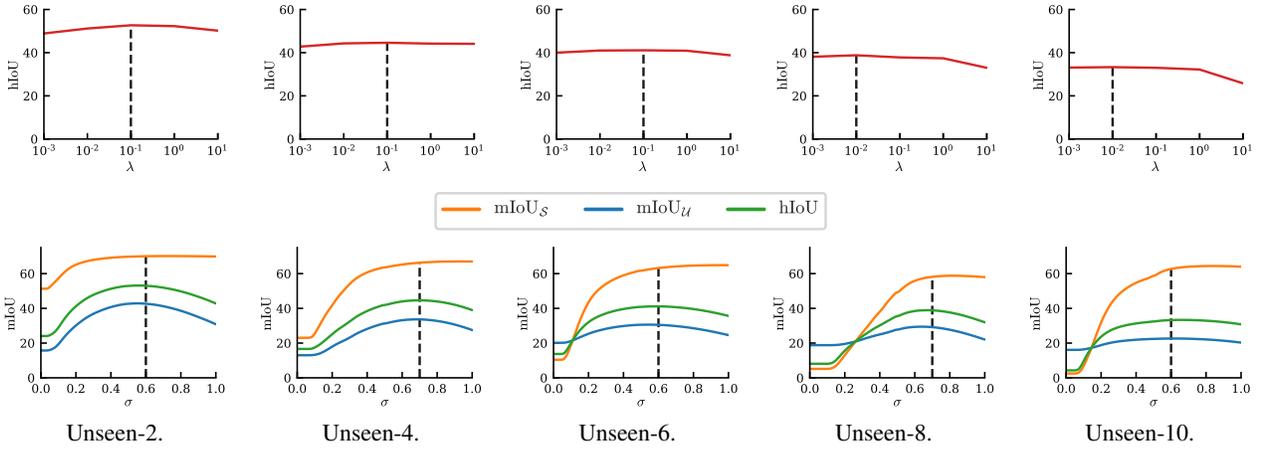

Figure 3. Comparison of mIoU scores for different hyperparameters $\lambda$ (top) and $\sigma$ (bottom) on each split of PASCAL VOC [6]. The dotted line indicates the value used in the main paper.

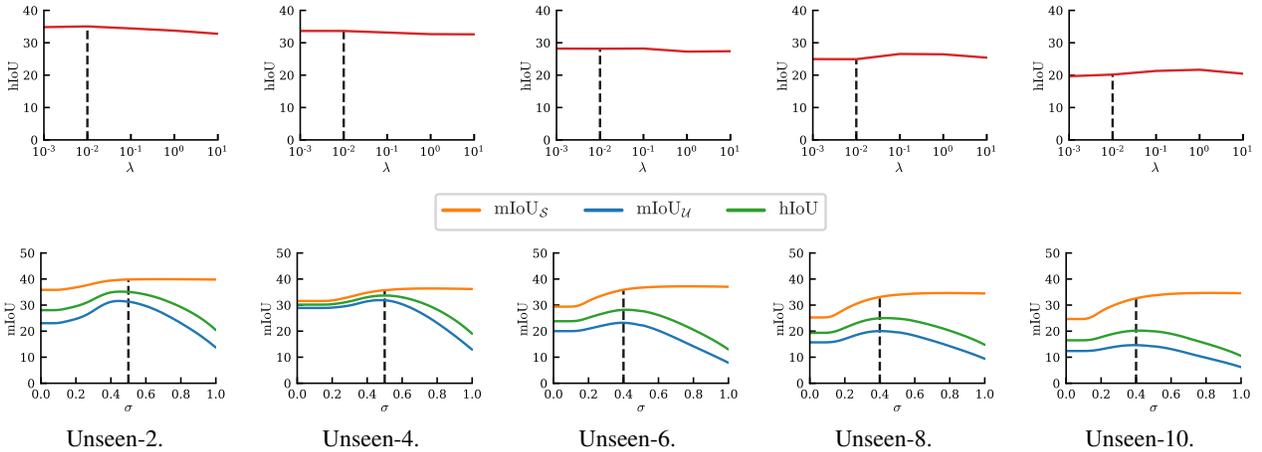

Figure 4. Comparison of mIoU scores for different hyperparameters $\lambda$ (top) and $\sigma$ (bottom) on each split of PASCAL Context [11]. The dotted line indicates the value used in the main paper.

Table 2. Per-class mIoU scores on the unseen-4 split of PASCAL VOC [6]. *: unseen classes; †: reimplementation; ‡: our visual encoder. The results of ZS3Net [2] are obtained from the model provided by the authors, but differ from the original ones. Numbers in bold are the best performance and underlined one is the second best.

| Method | bg. | aero* | bike | bird | boat | bot | bus | car | cat | cha | cow* | tab | dog | hor | mbik* | pers | plnt | she | sofa* | trai | tv | hIoU |
|---|---|---|---|---|---|---|---|---|---|---|---|---|---|---|---|---|---|---|---|---|---|---|
| ZS3Net [2] | **92.2** | 32.8 | **35.7** | 85.2 | 56.1 | **78.9** | **91.0** | 77.5 | 84.3 | **26.2** | 27.3 | 41.9 | 80.4 | **57.2** | 39.5 | **81.5** | **58.8** | 0.0 | 9.1 | 79.5 | **69.3** | 38.3 |
| ZS3Net† | 91.8 | **37.1** | 32.2 | **85.8** | 57.2 | 75.8 | **91.5** | 71.8 | **89.8** | 24.8 | **30.4** | 49.8 | **86.6** | 51.7 | 43.0 | 81.1 | 55.6 | 72.2 | 4.5 | 84.2 | 67.7 | 40.6 |
| ZS3Net‡ | 91.0 | 35.6 | 32.1 | 85.5 | 58.3 | 73.9 | 90.0 | **77.6** | 86.4 | 25.0 | **31.7** | 53.9 | 83.1 | 53.0 | 48.3 | 80.5 | 51.5 | 73.0 | 11.7 | 85.3 | 62.4 | 43.4 |
| Ours | 90.2 | 36.2 | 29.6 | 83.5 | **60.4** | 69.7 | 90.3 | 76.8 | 87.8 | 18.6 | 30.1 | **49.8** | 83.8 | 52.1 | **52.9** | 78.1 | 51.2 | 58.4 | **15.3** | **84.8** | 62.2 | **44.6** |

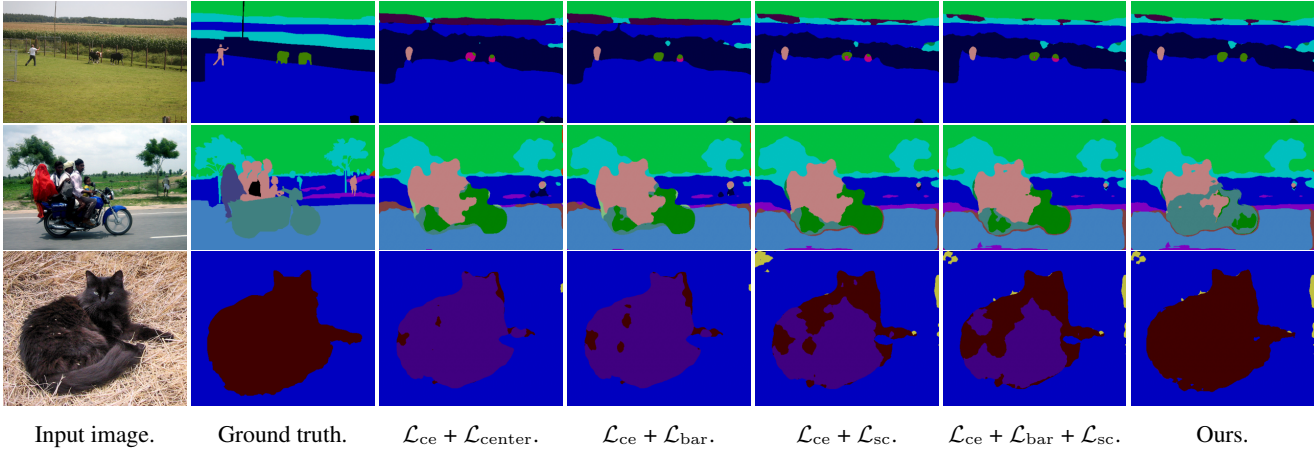

| Input image. | Ground truth. | $\mathcal{L}_{ce} + \mathcal{L}_{center}$. | $\mathcal{L}_{ce} + \mathcal{L}_{bar}$. | $\mathcal{L}_{ce} + \mathcal{L}_{sc}$. | $\mathcal{L}_{ce} + \mathcal{L}_{bar} + \mathcal{L}_{sc}$. | Ours. |

Figure 5. Visual comparison using different losses in our framework on the unseen-4 split of PASCAL Context [11]. Note that cow, motorbike, and cat are unseen classes.

## 2.2. SPNet.

To further demonstrate the effectiveness of our approach, we follow the experiment setting provided by SPNet [12] on PASCAL VOC [6]. Specifically, it provides a single split that consists of 15 seen and 5 unseen classes (potted-plant, sheep, sofa, train, and tv). For fair comparison, we use DeepLabV2 [3] with ResNet-101 [8] as our visual encoder. ResNet-101 is initialized by pre-trained weights for ImageNet classification [5]. For side information, we concatenate *word2vec* [10] and *fastText* [9], resulting in a 600-dimensional semantic feature. We use the same training details aforementioned in the main paper to train both encoders.

We have found that many virtual prototypes are located at boundaries between foreground and background. As the setting provided by SPNet ignores background regions, we set $r$ to 2. For temperature parameters, we adopt the same values as in the experimental settings provided by ZS3Net. Other hyperparameters ($\lambda = 1, \sigma = 0.8$) are chosen by a cross-validation as in [1]. Fig. 6 shows performance variations w.r.t. the value of $r$ and $\sigma$. We can see that the behavior of mIoU scores w.r.t. these hyperparameters is similar to the ones in Figs. 3 and 4.

We compare in Table 3 our approach with state-of-the-art GZS3 methods [2, 7, 12]. All numbers for other methods are taken from CaGNet [7]. From this table, we can clearly see that our approach outperforms all other methods including generative methods [2, 7] by a large margin in terms of mIoU$_\mathcal{U}$ and hIoU. Note that our approach without using AC already outperforms all other methods, demonstrating the effectiveness of the discriminative approach. In Fig. 9, we show qualitative comparison of ours and CaGNet. We can clearly see that our approach provides better results.

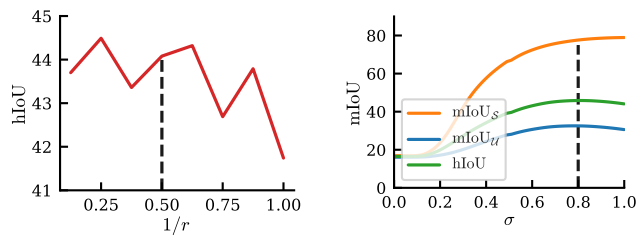

Figure 6. Comparison of mIoU scores for different hyperparameters $r$ (left) and $\sigma$ (right) on the experiment setting provided by SPNet [12]. We empirically set $\lambda$ to 1. The dotted line indicates the value chosen by a cross-validation.

Table 3. Quantitative comparison of state-of-the-art methods [2, 7, 12] and ours on the experiment setting provided by SPNet [12] in terms of mIoU.

| Methods | mIoU$_\mathcal{S}$ | mIoU$_\mathcal{U}$ | hIoU |
|---|---|---|---|
| SPNet [12] | 78.0 | 15.6 | 26.1 |
| ZS3Net [2] | 77.3 | 17.7 | 28.7 |
| CaGNet [7] | 78.4 | 26.6 | 39.7 |
| Ours w/o AC | 78.9 | 30.6 | 44.1 |
| Ours | 77.7 | 32.5 | **45.9** |

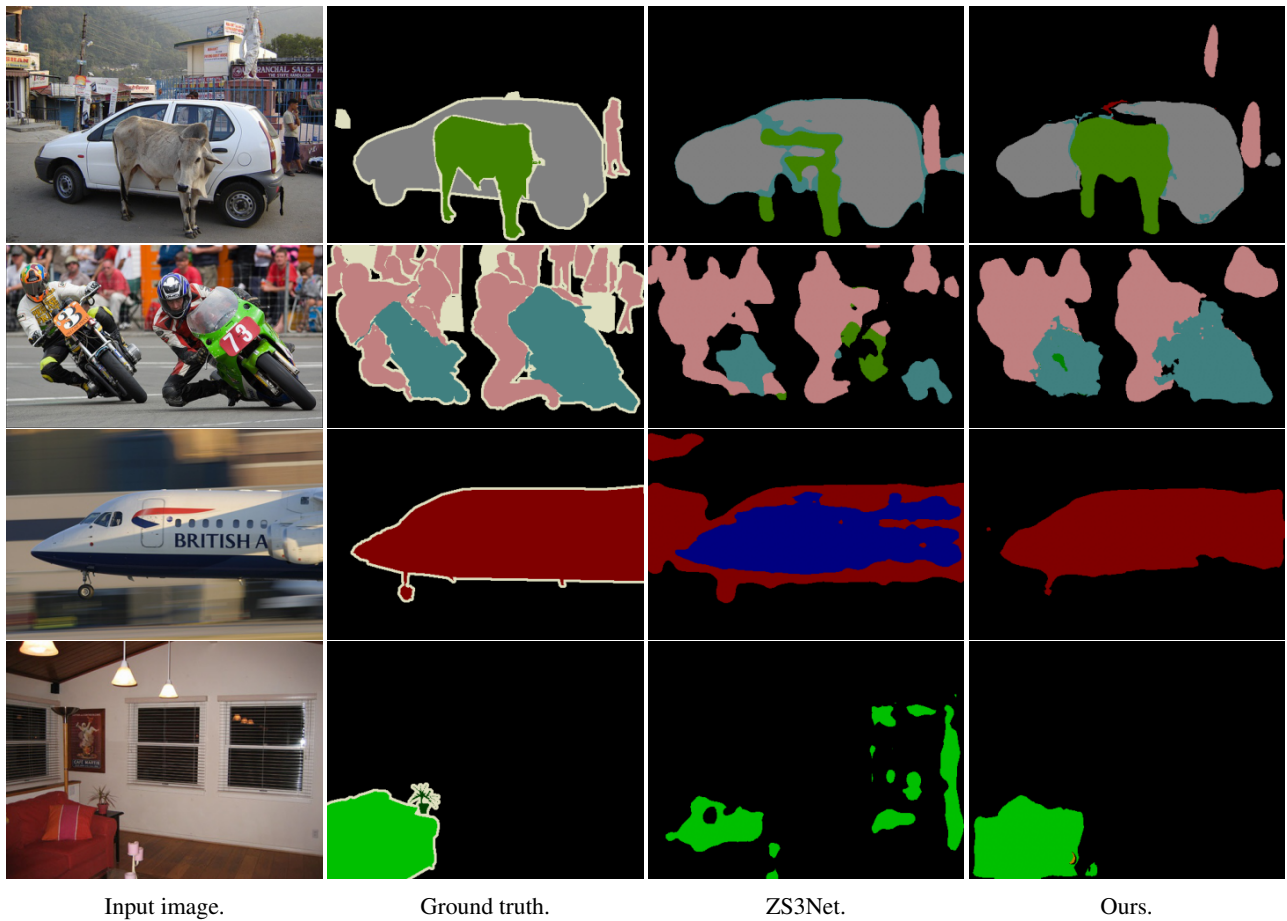

| Input image. | Ground truth. | ZS3Net. | Ours. |

Figure 7. Visual comparison of ZS3Net [2] and ours on the unseen-4 split of PASCAL VOC [6]. Note that cow, motorbike, airplane, and sofa are unseen classes.

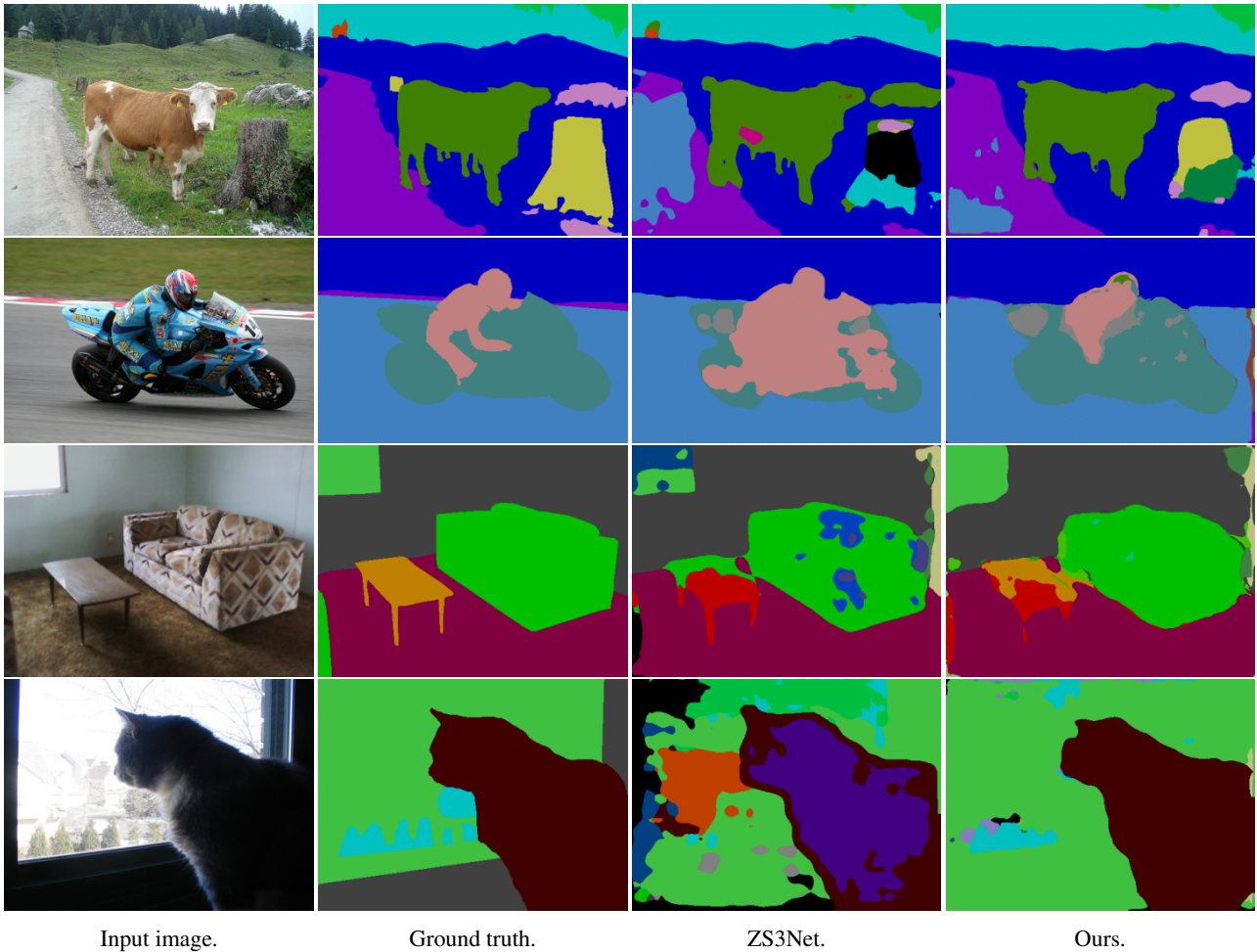

| Input image. | Ground truth. | ZS3Net. | Ours. |

Figure 8. Visual comparison of ZS3Net [2] and ours on the unseen-4 split of PASCAL Context [11]. Note that cow, motorbike, sofa, and cat are unseen classes.

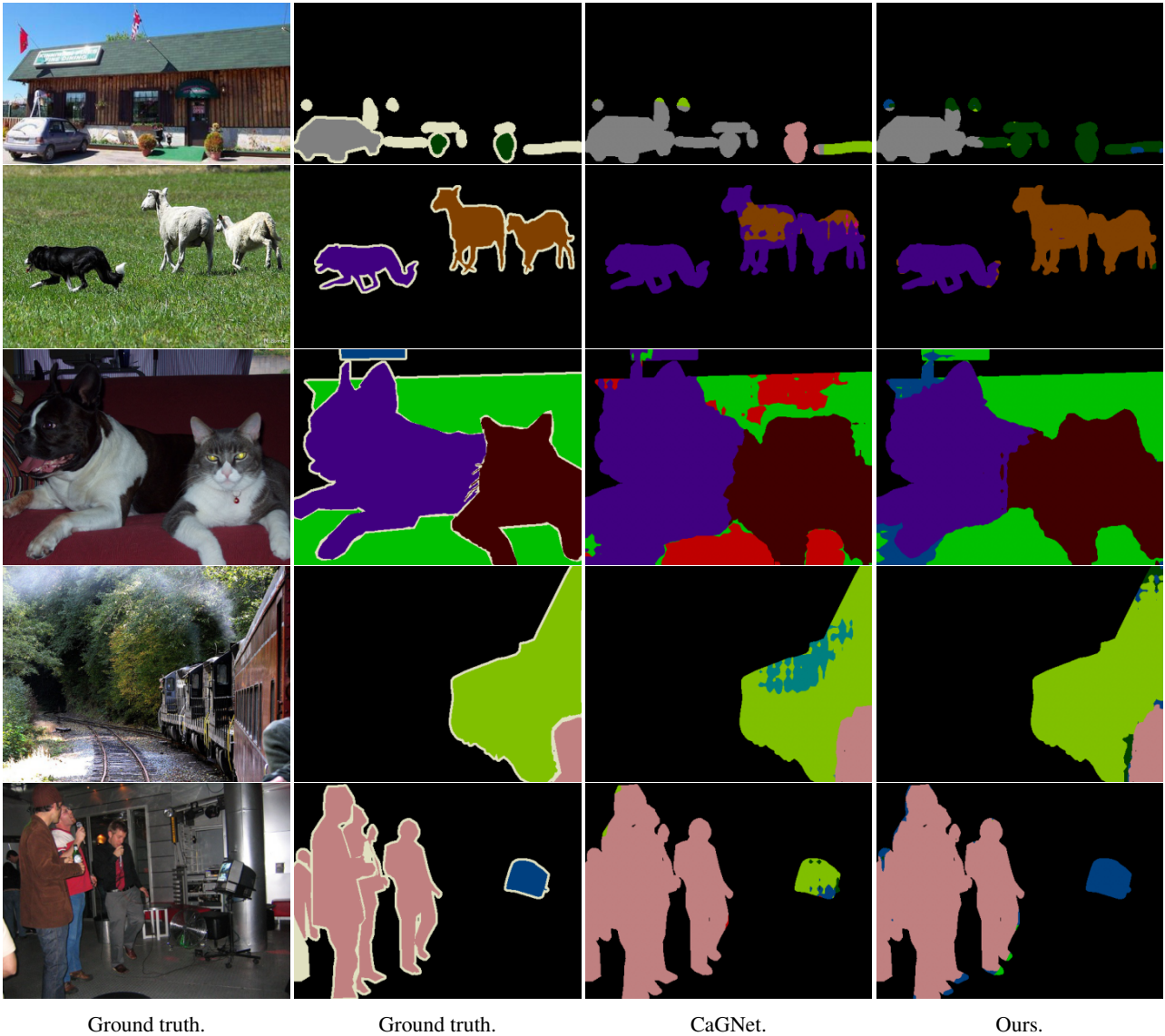

Figure 9. Visual comparison of CaGNet [7] and ours on the experiment setting provided by SPNet [12], where potted-plant, sheep, sofa, train, and tv are unseen classes. Note that all results show sharp object boundaries, since this setting uses pixel-wise annotations for the background class during inference.



\end{document}